\documentclass[12pt,journal,compsoc,onecolumn]{IEEEtran}
 \usepackage{amsmath,graphicx}
\usepackage{color}
 \usepackage{amssymb}
\usepackage{amsfonts}
\usepackage{dsfont}
\usepackage{epsfig}
 \usepackage{times}
\usepackage{graphicx}
\usepackage{adjustbox}
\usepackage{slashbox}
\usepackage{nicefrac}       
\usepackage{microtype}       
\usepackage{graphicx}
\usepackage{multirow}

 \usepackage{pifont}
\usepackage{lipsum}

 \usepackage{bbm}
 \usepackage{algorithmic}
\usepackage[noadjust]{cite}

\newcommand{\xmark}{\ding{55}}

\ifCLASSINFOpdf
\else
\fi
\ifCLASSOPTIONcompsoc
 \usepackage[font=normalsize,labelfont=sf,textfont=sf]{subfig}
\else
 \usepackage[font=footnotesize]{subfig}
\fi
\ifCLASSOPTIONcompsoc
 \usepackage[font=normalsize,labelfont=sf,textfont=sf]{subfig}
\else
 \usepackage[font=footnotesize]{subfig}
\fi

\begin{document}
\def\x{{\mathbf x}}
\def\L{{\cal L}}
\def\S{{\cal S}}
\def\V{{\cal V}}

\title{Deep hierarchical pooling design for cross-granularity action recognition} 
\author{Ahmed Mazari  \ \ \ \ \ \ \ \ \  \ \ \ Hichem Sahbi\\
Sorbonne University, CNRS, LIP6\\
F-75005, Paris, France 
}

\maketitle

\begin{abstract}
  In this paper, we introduce a novel hierarchical aggregation design that captures different levels of temporal granularity in action recognition. Our design principle is coarse- to-fine and achieved using a tree-structured network; as we traverse this network top-down, pooling operations are getting less invariant but timely more resolute and well localized. Learning the combination of operations in this network --- which best fits a given ground-truth —-- is obtained by solving a constrained minimization problem whose solution corresponds to the distribution of weights that capture the contribution of each level (and thereby temporal granularity) in the global hierarchical pooling process. Besides being principled and well grounded, the proposed hierarchical pooling is also video-length agnostic and resilient to misalignments in actions. Extensive experiments conducted on the challenging UCF-101 database corroborate these statements.
\end{abstract}
\hspace{1cm} {\small \noindent  {\bf  Keywords---} Hierarchical pooling, deep multiple representation learning, action recognition}
 
 \section{Introduction}
\label{sec:intro}

Many applications such as video surveillance~\cite{sahbiicip09,video_surv01,sahbiigarss12b,video_surv02,sahbiigarss12,sahbiicpr18,sahbijstars17,icip2001}, scene captioning and understanding  \cite{video_capt01,video_capt02,video_capt03,sahbispie2004,video_capt04,video_capt05,sahbiclef08,scene01,scene02,scene03,sahbiijmir15,GG-CNN,HC-MTL,sahbikpca06,action_localization04,MVSV,sahbijmlr06,sahbiaccv2010,sahbiclef13,sahbisc7,sahbicassp11} as well as robotics \cite{vid_robotics01,vid_robotics02,vid_robotics03,vid_robotics04,vid_robotics05} require automatic recognition of human actions. This task is one of the most challenging problems in video analysis which consists in assigning action categories to image sequences. The difficulty of this task stems from the intrinsic properties of actions (human appearance and motion, articulation, velocity, etc.) and also their extrinsic acquisition conditions (camera motion and resolution, illumination, occlusion, cluttered background, etc). Existing action recognition solutions  process videos in order to extract (handcrafted or learned) representations~\cite{hog,of,stips,lingsahbi2013,bag_features,Fisher,handcrafted_and_learned,phong,tnnls19} prior to their classification using shallow~\cite{temporalpyramid_detec,mkl_action,lingsahbieccv2014,lingsahbiicip,temporal_pyramid,lingsahbi2013,superived_dic_action,multi_svm,sahbipr2012,lingsahbiicip2014,sahbicvpr08a,linear_svm,sahbiphd,sahbifleuret02,rbf,polynomial_kernel,fuzzy05,icml08} or deep models~\cite{twostream14,spresnet16,mklimage2017,spresnetmulti17,temporalpyramid,pose,sahbiicassp16b,sahbiiccv17}. The latter are particularly powerful in visual recognition \cite{image_class01,image_class02} (and other neighboring fields~\cite{speech_reco01,speech_reco02}) and successful methods include 2D/3D two-stream convolutional neural networks (CNNs) \cite{twostream14,segment_net}. This success, which comes at the expense of a substantial increase in the number of training parameters, is tributary to the availability of large {\it labeled} video datasets that capture all the intrinsic and the extrinsic properties of scenes and actions. However, labeled videos are scarce and existing ones are at least an order of magnitude smaller compared to the datasets used in other related tasks (such as image classification) while action recognition is inherently far more challenging. As a result, deep networks used for action recognition become more exposed to over-fitting. \\  

\indent Deep convolutional networks have nonetheless the ability to attenuate the high dependency on labeled data by introducing pooling (a.k.a aggregation) operators which gradually reduce the dimensionality, the number of training parameters and thereby the risk of over-fitting. However, pooling (such as averaging) may dilute the relevant information especially when action categories exhibit strong variations in their temporal granularity. Indeed, while coarsely-grained actions could still remain easy to discriminate using average pooling, fine-grained ones become more confound; hence, one should design a pooling mechanism which conveys multiple levels of granularity across categories. \\

\indent In this paper, we introduce a novel hierarchical pooling (aggregation) design that captures different levels of temporal granularity in action recognition. Our design principle is ``coarse-to-fine'' and achieved using a tree-structured network; as we parse this hierarchy top-down, pooling operations are getting less invariant but timely more resolute and well dedicated to fine-grained action categories. Given a hierarchy of aggregation operations, our goal is to learn weighted (linear and nonlinear) combinations of these pooling operations that best fit a given action recognition ground-truth. We solve this problem by minimizing a constrained objective function whose parameters correspond to the distribution of weights through multiple aggregation levels; each weight measures the contribution of its granularity in the global learned video representation. Besides being able to handle aggregations at different levels, the particularity of our solution resides in its ability to handle {misaligned\footnote{\scriptsize misalignments are usually due to imprecise detection and trimming of actions in videos (which is also known to be a cumbersome task when achieved manually and error-prone when achieved automatically~\cite{action_localization01,action_localization02,action_localization03,action_localization05}) and this adds spurious details/context in the analyzed actions.} and variable duration} videos (without any explicit alignment or up/down-sampling) and thereby makes it possible to fully benefit from the whole frames in videos. Extensive experiments conducted on the challenging UCF-101 benchmark show the validity and the out-performance of our hierarchical aggregation design w.r.t the related work. 

\section{Frame-wise  description at a Glance}\label{approach1}
We consider a collection of videos $\S = \{\V_i \}_{i=1}^n$ with each one being a sequence of frames $\V_i=\{f_{i,t}\}_{t=1}^{T_i}$  and a set of action categories (a.k.a classes or categories) denoted as ${\cal C} = \{1,\dots, C\}$. In order to describe the visual content of a given video $\V_i$, we rely on a two-stream process; the latter provides a complete description of appearance and motion that characterizes the spatio-temporal aspects of moving objects and their interactions. The output of the appearance stream (denoted as $\{\phi_a(f_{i,t})\}_{t=1}^{T_i} \subset \mathbb{R}^{2048}$) is based on the deep residual network (ResNet-101) trained on ImageNet \cite{pretrained_resnet101_ucf} and fine-tuned on UCF-101 while the output of the motion stream (denoted as $\{\phi_m(f_{i,t})\}_{t=1}^{T_i} \subset \mathbb{R}^{2048}$) is also based on  ResNet-101 but trained on optical flow input frames \cite{pretrained_resnet101_ucf,segment_net};  in the appearance stream, the number of input channels in the underlying ResNet is kept fixed (equal to $3$) while in the motion stream, the number of channels is reset to $20$ (instead of $3$).
 
When training the latter, the initial weights of these 20 channels are obtained by averaging the 3 original (appearance) channel weights and by replicating their values through the 20 new motion channels. Considering these frame-wise representations, our goal is to introduce an alternative to usual frame aggregation schemes (namely sampling and global average pooling) which instead learns a hierarchical aggregation that models  {\it coarse as well as fine grained action categories}.

 \def\N{{\cal N}}
 \def\T{{\cal T}}
 \def\betaa{{\bf \beta}}
 \def\w{{\bf w}}                 

 \section{Multiple Aggregation Learning}\label{approach2}
 
\indent Given a video $\V$ with $T$ frames, we define $\N$ as a tree-structured network with depth up to $D$ levels and width up to $2^{D-1}$. Let $\N = \cup_{k,l} \N_{k,l}$ with $\N_{k,l}$ being the $k^{th}$ node of the $l^{th}$ level of $\N$; all nodes belonging to the $l^{th}$ level of $\N$ define a partition of the temporal domain $[0,T]$ into $2^{l-1}$ equally-sized subdomains. A given node $\N_{k,l}$ in this hierarchy aggregates the frames that belong to its underlying temporal interval. Each node $\N_{k,l}$ also defines an appearance and a motion representation respectively denoted as  $\psi^a_{k,l}(\V_i)$, $\psi^m_{k,l}(\V_i)$ and set as  $\psi^a_{k,l}(\V_i)=\frac{1}{|\N_{k,l}|}\sum_{t \in \N_{k,l}} \phi_a(f_{i,t})$, $\psi^m_{k,l}(\V_i)=\frac{1}{|\N_{k,l}|}\sum_{t \in \N_{k,l}} \phi_m(f_{i,t})$. Depending on the level in $\N$, each representation captures a particular temporal granularity of motion and appearance into a given scene; it is clear that top-level representations capture coarse visual characteristics of actions while bottom-levels (including leaves) are dedicated to fine-grained and timely-resolute sub-actions. Knowing a priori which levels (and nodes in these levels) capture the best -- a given action category  -- is not trivial. In the remainder of this section, we introduce a novel learning framework which achieves multiple aggregation design and finds the best combination of levels and nodes in these levels that fits  different temporal granularities of action categories. 
\subsection{Multiple aggregation learning} 
\indent Considering the motion stream, we define -- for each node $\N_{k,l}$ -- a set of variables $\beta_m=\{\betaa_{k,l}^m\}_{k,l}$ (with $\betaa_{k,l}^m \in [0,1]$ and $\sum_{k,l} \betaa_{k,l}^m= 1$) which measures the importance (and hence the contribution) of $\psi^m_{k,l}(\V)$ in the global motion representation of $\V$ (denoted as $\psi_m(\V)$). Precisely, two variants are considered for $\psi_m$ 
\begin{equation}\label{eq1}
  \begin{array}{lll}
   \textrm{(*)}  &  \psi_m(\V) & = \displaystyle \left(\betaa_{1,1}^m \psi_{1,1}^m(\V)  \dots  \betaa_{k,l}^m \psi_{k,l}^m(\V) \dots \right)^\top \\
                 &  & \\
   \textrm{(**)}   & \psi_m(\V) &= \displaystyle \sum_{k,l} \betaa_{k,l}^m \psi^m_{k,l}(\V).  
                     \end{array}
                   \end{equation}                
                   \noindent As shown  above, the variant in (*) corresponds to a concatenation scheme while (**) corresponds to averaging; the {\it former} relies on the hypothesis that nodes in ${\cal N}$ (and hence sub-actions in different videos) are well aligned whereas the {\it latter} relaxes this hypothesis (see later Eq.~\ref{eq0000}).  Similarly to motion, we define the aggregations and the set of variables $\betaa_a=\{\betaa_{k,l}^a\}_{k,l}$  associated to appearance stream. In the remainder of this paper, and unless explicitly mentioned, the symbols $m$, $a$ are omitted in the notation and all the subsequent formulation is applicable to motion as well as appearance streams.\\
                   
\def\K{{\cal K}}
\indent In order to weight the impact of nodes in the hierarchy $\cal N$ and put more emphasis on the most relevant granularity of the learned aggregation, we consider multiple representation learning that generalizes~\cite{MKL,MKL_alignement}  both to linear and nonlinear combinations. Its main idea consists in finding a kernel $\K$ as a combination of positive semi-definite (p.s.d) elementary kernels $\{\kappa(.,.)\}$ associated to $\{\N_{k,l}\}_{k,l}$. Considering the two maps in Eq.~(\ref{eq1}), we define the two variants of $\cal K$ as
\begin{equation} 
  \begin{array}{lll}   
    \K(\V,\V') &=&\displaystyle \sum_{l} \sum_{k}\beta_{k,l} \ \kappa(\psi_{k,l}(\V),\psi_{k,l}(\V')) \\
                  \K(\V,\V') &=&\displaystyle \sum_{l,l'} \sum_{k,k'}\beta_{k,l} \beta_{k',l'} \ \kappa(\psi_{k,l}(\V),\psi_{k',l'}(\V')).
                   \end{array}\label{eq0000}
\end{equation}
As $\beta_{k,l} \in [0,1]$, the kernel $\K$ is p.s.d resulting from the closure of the p.s.d of $\kappa$ w.r.t the sum and the product. Let ${\cal C}=\{1,\dots,C\}$ be a set of action categories and let $\{(\V_i,y_{ic})\}_i$ be a training set of actions associated to $c \in {\cal C}$ with $y_{ic} =+1$ if $\V_i$ belongs to the category $c$ and $y_{ic} =-1$ otherwise. Using $\K$, we train multiple max margin classifiers (denoted $\{g_c\}_{c \in {\cal C}}$) whose kernels (in Eq.~\ref{eq0000}) correspond to level-wise linear (resp. cross-wise nonlinear) combinations of elementary kernels dedicated to $\{{\cal N}_{k,l}\}_{k,l}$. A classifier associated to an action category $c$ is given by $g_c(\V) = \sum_{i} \alpha_i^c  y_{ic}  \K(\V,\V_i) + b_c$, here $b_c$ is a shift, $\{\alpha_i^c\}_i$ is a set of positive parameters found (together with $\beta=\{\beta_{k,l}\}_{k,l}$) by minimizing the following constrained quadratic programming (QP) problem
\begin{equation}
  \begin{array}{ll}
    \displaystyle 
    \min_{0 \leq \beta \leq 1, \|\beta \|_1=1,\{\alpha_i^c\}} &  \displaystyle  \frac{1}{2} \sum_c \sum_{i,j} \alpha_{i}^c \alpha_j^c y_{ic} y_{jc}  \K(\V_i,\V_j) - \sum_i \alpha_i^c \\
    \textrm{s.t.}              &   \displaystyle \alpha_{i}^c\geq 0,  \ \ \ \ \ \  \sum_{i} y_{ic} \alpha_{i}^c=0, \ \ \  \forall i, c. 
\end{array}\label{optimiz}
\end{equation}
As the problem  in Eq.~\ref{optimiz} is not convex w.r.t $\beta$, $\{\alpha_i^c\}$ taken jointly and convex when taken separately, an EM-like iterative optimization procedure can be used: first, parameters in $\beta$ are fixed and the above problem is solved w.r.t $\{\alpha_i^c\}$ using QP, then  $\{\alpha_i^c\}$ are fixed and the resulting problem is solved w.r.t  $\beta$ using either linear programming for (*)  and QP for (**). This iterative process stops when the values of all these parameters remain unchanged or when it reaches a maximum number of iterations. However, in spite of being relatively effective (see later Table~\ref{tab:deep_MKL}), this EM-like procedure is computationally expensive as it requires solving multiple instances of constrained quadratic problems\footnote{\scriptsize whose complexity scales quadratically w.r.t the size of training data and the number of nodes in the hierarchy $\cal N$.} and the number of necessary iterations to reach convergence is large in practice.
\begin{figure}[h!]
    \centering
     \includegraphics[width=0.49\columnwidth]{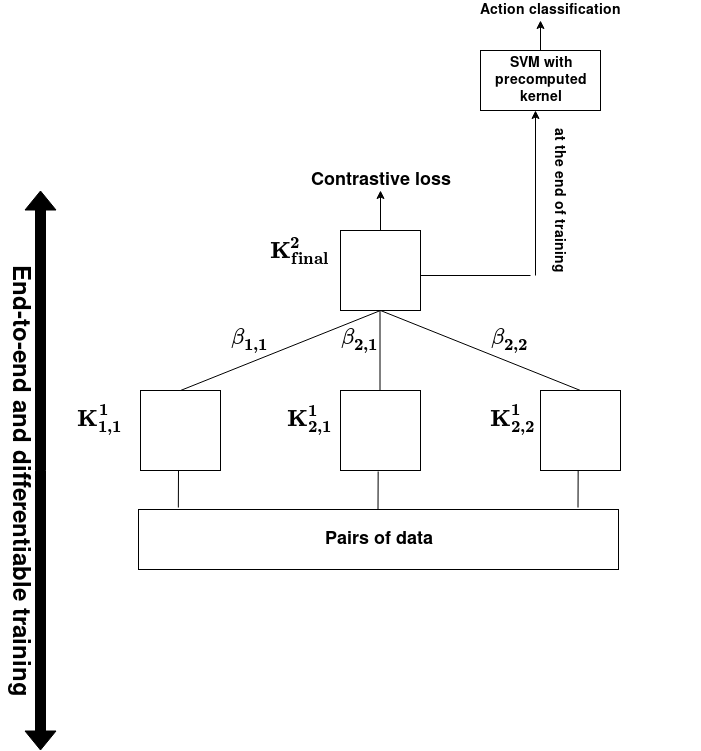}  \includegraphics[width=0.49\columnwidth]{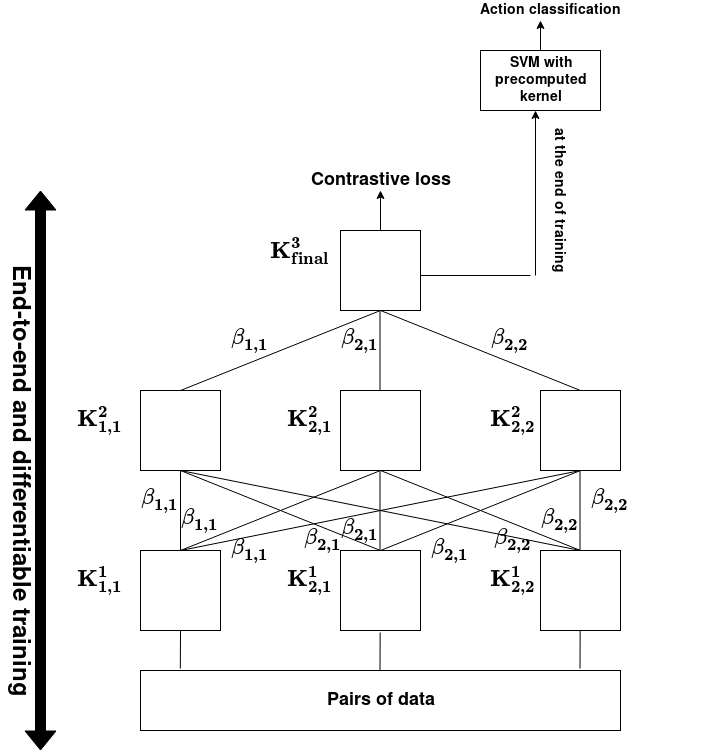}
    \caption{Examples of networks used to train a 2-level hierarchical aggregation with ``concatenation'' (left) and ``averaging'' (right). These two networks correspond to the two equations in (\ref{eq0000}); their inputs correspond to the elementary kernels evaluated on pairwise nodes in ${\cal N}$: constrained to be aligned for ``concatenation'' and unconstrained for ``averaging'' {\bf (Better to zoom the PDF version).}}
    \label{fig:model}
\end{figure}
\subsection{Deep contrastive loss design}
In what follows, we consider a procedure that decouples the learning of $\beta$ from $\{\alpha_i^c\}$ resulting into more efficient and also still effective training process. In this procedure, we first model the kernels in Eq.~\ref{eq0000} using two networks (see Fig.~\ref{fig:model}), and  we learn their parameters using a contrastive loss criterion (that benefits from larger training data pairs),  then we plug the resulting $\K$ into Eq.~\ref{optimiz} in order to learn the parameters $\{\alpha_i^c\}$ in one step. We consider an end-to-end framework which learns the parameters $\betaa$ of these networks (that capture the importance of nodes in the hierarchical aggregation) by minimizing
\begin{equation}\label{eq0} 
  \begin{array}{ll}
\displaystyle   \min_{0 \leq \betaa \leq 1, \| \beta \|_1 = 1} &  E(\beta,{\cal S},{\cal K},Y),
 \end{array} 
\end{equation}
\def\betaaa{{\hat{\beta}}}
\noindent here $E$ models the disagreement between the predicted kernel values on video pairs  $\{\K(\V_i,\V_j)\}_{\V_i,\V_j \in {\cal S}} $ and their ground-truth $\{Y(\V_i,\V_j)\}_{\V_i,\V_j \in {\cal S}}$ with $Y(\V_i,\V_j)=+1$ iff $\V_i$ and $\V_j$ belong to the same class and  $Y(\V_i,\V_j)=-1$ otherwise. This objective function can still be minimized using gradient descent and back-propagation. However, some constraints should be carefully tackled; indeed, whereas the forward step can be achieved, gradient back-propagation (through our multiple aggregation shown mainly in Fig.~\ref{fig:model}-right) should be achieved while sharing parameters in the same layers and across layers. Besides, constraints on $\beta's$ should also be handled. 
\subsection{Constraint implementation} 
Considering $\frac{\partial E}{\partial \K}$ available, the gradients $\frac{\partial E}{\partial \beta}$ cannot be straightforwardly obtained using a  direct application of the chain rule (as already available in PyTorch); on the one hand, any step following the gradient $\frac{\partial E}{\partial \beta}$ should preserve equality and inequality constraints in Eq.~(\ref{eq0}) while a direct application of the chain rule provides us with a surrogate gradient which ignores these constraints. On the other hand, as the parameters $\beta$ are shared across layers (when using ``averaging'' in Fig.~\ref{fig:model}), this requires a careful update of $\frac{\partial E}{\partial \beta}$ as shown subsequently.\\
\indent In order to implement the equality  and inequality constraints in Eq.~\ref{eq0}, we consider a re-parametrization as $\beta_{k,l}= h(\betaaa_{k,l})\slash {\sum_{k',l'} h(\betaaa_{k',l'})}$ for some $\{\betaaa_{k,l}\}_{k,l}$ with $h$ being strictly monotonic real-valued (positive) function and this allows free settings of the parameters  $\{\betaaa_{k,l}\}_{k,l}$ during optimization while guaranteeing $\beta_{k,l} \in [0,1]$ and $\sum_{k,l}  \beta_{k,l}=1$. During back-propagation, the gradient of the loss $E$ (now w.r.t $\betaaa$'s) is updated using the chain rule as
\begin{equation}
  \begin{array}{lll}
 \displaystyle   \frac{\partial E}{\partial \betaaa_{k,l}} &=& \displaystyle \sum_{p,q} \frac{\partial E}{\partial \beta_{p,q}} . \frac{\partial \beta_{p,q}}{\partial \betaaa_{k,l}} \\ 
              \ \ \ \ \textrm{with}   & & \ \ \ \displaystyle  \frac{\partial \beta_{p,q}}{\partial \betaaa_{k,l}} =   \displaystyle  \frac{h'(\betaaa_{k,l})}{\sum_{k',l'} h( \betaaa_{k',l'})} .  (\delta_{p,q,k,l}-\beta_{p,q}),                                                                                                                                                                                   \end{array}
                                                                                                                                                                                           \end{equation} 
                                                                                                                                                                                           \noindent and $\delta_{p,q,k,l}=1_{\{(p,q)=(k,l)\}}$. In practice $h(.)=\exp(.)$ and   $\frac{\partial E}{\partial \beta_{p,q}}$ is obtained from layerwise gradient backpropagation (as already integrated in standard deep learning tools including PyTorch).  Hence,  $\frac{\partial E}{\partial \betaaa_{k,l}}$ is obtained by multiplying the original gradient $\big[\frac{\partial E}{\partial {\beta}_{p,q}}\big]_{p,q}$ by the Jacobian  $\big[\frac{\partial \beta_{p,q}}{\partial \betaaa_{k,l}}\big]_{p,q,k,l}$ which simply reduces to  $\big[\beta_{k,l} (\delta_{p,q,k,l}-\beta_{p,q})\big]_{p,q,k,l}$ when  $h(.)=\exp(.)$. \\

                                                                                                                                                                                           \noindent  As the parameters $\{\betaaa_{k,l}\}_{k,l}$ are not totally independent across layers (see again Fig.~\ref{fig:model}-right), we consider a further step that accumulates (averages) the gradients  $\{\frac{\partial E}{\partial \betaaa_{k,l}}\}_{k,l}$ with shared indices and replaces these gradients by the averaged ones. It is easy to see that these accumulated (shared) gradients (when used to update $\betaaa$'s using gradient descent) also preserve the equality and inequality constraints in Eq.~\ref{eq0}.  
                                                                                                                                                                                           
\section{Experiments}\label{exp}

We evaluate the performance of our action recognition method on the challenging UCF-101 (split-2) dataset \cite{ucf}. The latter includes 13,320 videos belonging to 101 action categories of variable duration, cluttered background and misaligned content\footnote{\scriptsize Many actions are misaligned as their videos are endowed with large context while others are precisely trimmed and contain only the actions of interest}. As discussed previously, we first extract 2D two-stream frame-wise representations, then we combine them using our hierarchical aggregation design prior to achieve action recognition. We follow the exact protocol in \cite{ucf} to evaluate and compare our method w.r.t different settings as well as the related work.\\

\noindent{\bf Settings.} Different settings are considered in order to assess the performance of our method: i) multiple depths of our hierarchical aggregation network ranging from 2 to 6, ii) two streams (motion and appearance) as well as their fusion, and iii) the two types of aggregations namely  ``concatenation'' and ``averaging''. In order to learn the weights of our hierarchical aggregations for all the aforementioned settings, we conducted  experiments using both the EM-like procedure as well as the deep multiple kernel learning (DMKL) shown in section~\ref{approach2}. In the latter, we achieve DMKL for 4,000 iterations using PyTorch Adam optimizer\footnote{\scriptsize We run experiments on single GPU; GeForce RTX 2080 Ti (with 11 GB).} and we set the learning rate to 0.0005 and the batch-size to 2048. As already discussed, we use a contrastive loss for DMKL and we plug the resulting kernel into multi-class SVMs for training and testing; given a test video,  its category corresponds to the SVM with the highest score.\\

 \begin{table}[h!]
    \centering
     \resizebox{0.55\columnwidth}{!}{
    \begin{tabular}{ccccc}
  
          & \bf  Depth (D) & \bf Appearance & \bf Motion & \bf Fusion  \\
        \hline
         \multirow{5}*{\rotatebox[origin=c]{90}{\bf Concatenat.}}   
         &  2 & 82.78 & 80.12  & 89.49   \\
          &  3  & 82.91 & 80.59 & 89.68  \\
          &  4 & 83.04 & 80.73 &  89.72 \\
          &  5  & {\bf 83.17} & {\bf 80.80}  & {\bf 89.76}  \\
          &  6  & 82.76  & 80.62 & 89.63 \\
          \hline
          \hline
        \multirow{5}*{\rotatebox[origin=c]{90}{\bf Averaging}}  
         &  2 &82.96 & 80.53 & 89.67   \\
          & 3  & 83.16 & 80.78 & 89.74  \\
          & 4 & 83.28 & 81.00 &  89.87 \\
          & 5  & 83.36 & 81.00 &   89.89 \\
          & 6  & {\bf 83.36} & {\bf 81.07} & {\bf 89.91}\\

    \end{tabular}
    }
    \caption{This table shows the behavior of our multiple aggregation learning using the EM procedure w.r.t the depth of the network ${\cal N}$. These results are reported for both motion and appearance streams as well as their combination and also for concatenation and averaging (note that RBF is used as an elementary kernel for DMKL). The drop in the performances of the ``concatenation'' scheme (from $D=5$ to $D=6$, i.e., the most resolute nodes) is mainly due to the sensitivity of ``concatenation'' to misalignments in the most resolute nodes of $\cal N$ while ``averaging'' enhances the performances steadily.}
    \label{tab:deep_MKL}
\end{table}

\noindent{\bf Performances and comparison.} Table~\ref{tab:deep_MKL} shows the performances of the different configurations (described earlier); from these results, we observe a consistent gain as the depth of our hierarchy increases with an advantage of ``averaging'' w.r.t ``concatenation''. This gain is observed on both motion and appearance streams with a significant leap when fusing them. These gains also reflect the importance of node crossbreeding (``averaging'' vs. ``concatenation'') especially when videos are subject to cluttered context and when their actions are misaligned as frequently observed in the UCF-101. \\ 
\indent We also show (in Table.~\ref{tab:comparison}) a comparison of our hierarchical aggregation  against two other aggregation methods:  global average pooling and also spectrograms~\cite{temporalpyramid}; the {\it former} produces a global representation that averages all the frame descriptions while the {\it latter} keeps all the frame representations and concatenate them (as an image) prior to their classification using 2D CNNs~\cite{temporalpyramid}. Note that these two comparative methods are interesting as they correspond to two extreme cases of our hierarchy, namely the root and the leaf levels; in particular, the spectrogram (of a video $\V$ with $T$ frames) is obtained when the number of leaf nodes, in the hierarchy $\cal N$, is exactly equal to $T$~(see again \cite{temporalpyramid}). We also compare our  method against another aggregation method based on colorized heatmaps~\cite{pose} as a variant of the global average pooling; these heatmaps correspond to timely-stamped and averaged frame-wise probability distributions of human keypoints. Finally, we compare the classification performances of our method against two closely related 2D CNN action recognition works: 2D two-streams CNNs in \cite{twostream14} and  \cite{pretrained_resnet101_ucf} (respectively based on VGG and ResNet) as well as the method in~\cite{C3D}. From these results, we observe a consistent gain of our hierarchical aggregation design w.r.t these related methods.  

\begin{table}[h!]
    \centering
    \resizebox{0.65\columnwidth}{!}{
    \begin{tabular}{ccccc}
       & \bf Methods & \bf Appearance & \bf Motion & \bf Fusion  \\
    \hline
         & Our HA+C   (EM)    & 82.76 & 80.62  &  89.63  \\
                  & Our HA+C   (DMKL)    & 82.82 & 80.69 & 89.66  \\
         & Our HA+A   (EM)    & 83.36 & 81.07  & 89.91   \\
        & Our HA+A (DMKL) & {\bf  83.44} & 81.17  & {\bf 89.95} \\
             \hline
            \hline
                  & GAP  in \cite{temporalpyramid}    & 66.15  & \xmark  & \xmark  \\
        & Spectrogram  \cite{temporalpyramid}  & 64.41& \xmark & \xmark  \\
             & Colorized heatmaps \cite{pose} &\xmark  &64.38 &\xmark \\
       &C3D \cite{C3D}  & 82.3 & \xmark & \xmark \\
        &Temporal Pyramid \cite{temporalpyramid}        &68.58 & \xmark &  \xmark  \\
                &2D 2-stream VGG \cite{twostream14} & 73 & {\bf  83.7} & 86.9 \\
        &2D 2-stream ResNet \cite{pretrained_resnet101_ucf} &  82.1 &79.4& 88.5\\
    \end{tabular}}
    \caption{ Comparison w.r.t state-of-the-art methods. In this table HA, C, A, GAP stand for ``Hierarchical Aggregation'', ``Concatenation'', ``Averaging'' and ``Global Average Pooling'' respectively. Note that our HA results are obtained with $D=6$ and RBF is used for both EM and DMKL settings. In the related work, the symbol ``\xmark''  means that the configuration either ``does not apply'' or ``not tested'' in the related paper.}
    \label{tab:comparison}
  \end{table}
  
\section{Conclusion}
In this paper, we introduced a hierarchical aggregation design for cross-granularity action recognition. Our method is based on the minimization of a constrained objective function whose solution corresponds to the distribution of weights in a hierarchy of pooling operations that best fits the granularity of action categories. Besides being able to handle videos with multiple granularities, the strength of our  method resides also in its ability to handle videos with variable duration and misalignment. Experiments conducted on UCF-101 dataset show the validity of our approach {\it w.r.t} the related work. As future work, we are currently investigating the extension of our hierarchical crossbreeding aggregation method in order to handle longer videos as a part of the more challenging problem of activity recognition.

\end{document}